\def\year{2021}\relax
\documentclass[letterpaper]{article} % DO NOT CHANGE THIS
\usepackage{aaai21}  % DO NOT CHANGE THIS
\usepackage{times}  % DO NOT CHANGE THIS
\usepackage{helvet} % DO NOT CHANGE THIS
\usepackage{courier}  % DO NOT CHANGE THIS
\usepackage[hyphens]{url}  % DO NOT CHANGE THIS
\usepackage{graphicx} % DO NOT CHANGE THIS
\usepackage{natbib}  % DO NOT CHANGE THIS AND DO NOT ADD ANY OPTIONS TO IT
\usepackage{caption} % DO NOT CHANGE THIS AND DO NOT ADD ANY OPTIONS TO IT
\usepackage{amsmath}
\usepackage{array}
\usepackage{multirow}
\usepackage{booktabs}
\usepackage{arydshln}
\usepackage{xcolor}
\usepackage{amssymb}
\usepackage{bbm}

\DeclareMathOperator*{\argmin}{arg\,min}
\DeclareMathOperator*{\argmax}{arg\,max}
\newcolumntype{C}[1]{>{\centering\arraybackslash}m{#1}}

\title{Scene Graph Embeddings Using Relative Similarity Supervision}
\author {
    Paridhi Maheshwari\textsuperscript{\rm 1}\thanks{These authors contributed equally}, Ritwick Chaudhry\textsuperscript{\rm 2}\footnotemark[1]\thanks{Work done while at Adobe Research}, Vishwa Vinay\textsuperscript{\rm 1} \\
}
\affiliations {
    \textsuperscript{\rm 1} Adobe Research \\
    \textsuperscript{\rm 2} Carnegie Mellon University \\
    parimahe@adobe.com, rchaudhr@andrew.cmu.edu, vinay@adobe.com
}

\begin{document}

\maketitle

\begin{abstract}
Scene graphs are a powerful structured representation of the underlying content of images, and embeddings derived from them have been shown to be useful in multiple downstream tasks. In this work, we employ a graph convolutional network to exploit structure in scene graphs and produce image embeddings useful for semantic image retrieval. Different from classification-centric supervision traditionally available for learning image representations, we address the task of learning from relative similarity labels in a ranking context. Rooted within the contrastive learning paradigm, we propose a novel loss function that operates on pairs of similar and dissimilar images and imposes relative ordering between them in embedding space. We demonstrate that this Ranking loss, coupled with an intuitive triple sampling strategy, leads to robust representations that outperform well-known contrastive losses on the retrieval task. In addition, we provide qualitative evidence of how retrieved results that utilize structured scene information capture the global context of the scene, different from visual similarity search.
\end{abstract}

\section{Introduction}

\begin{figure*}[!h]
    \centering
    \includegraphics[width=\linewidth]{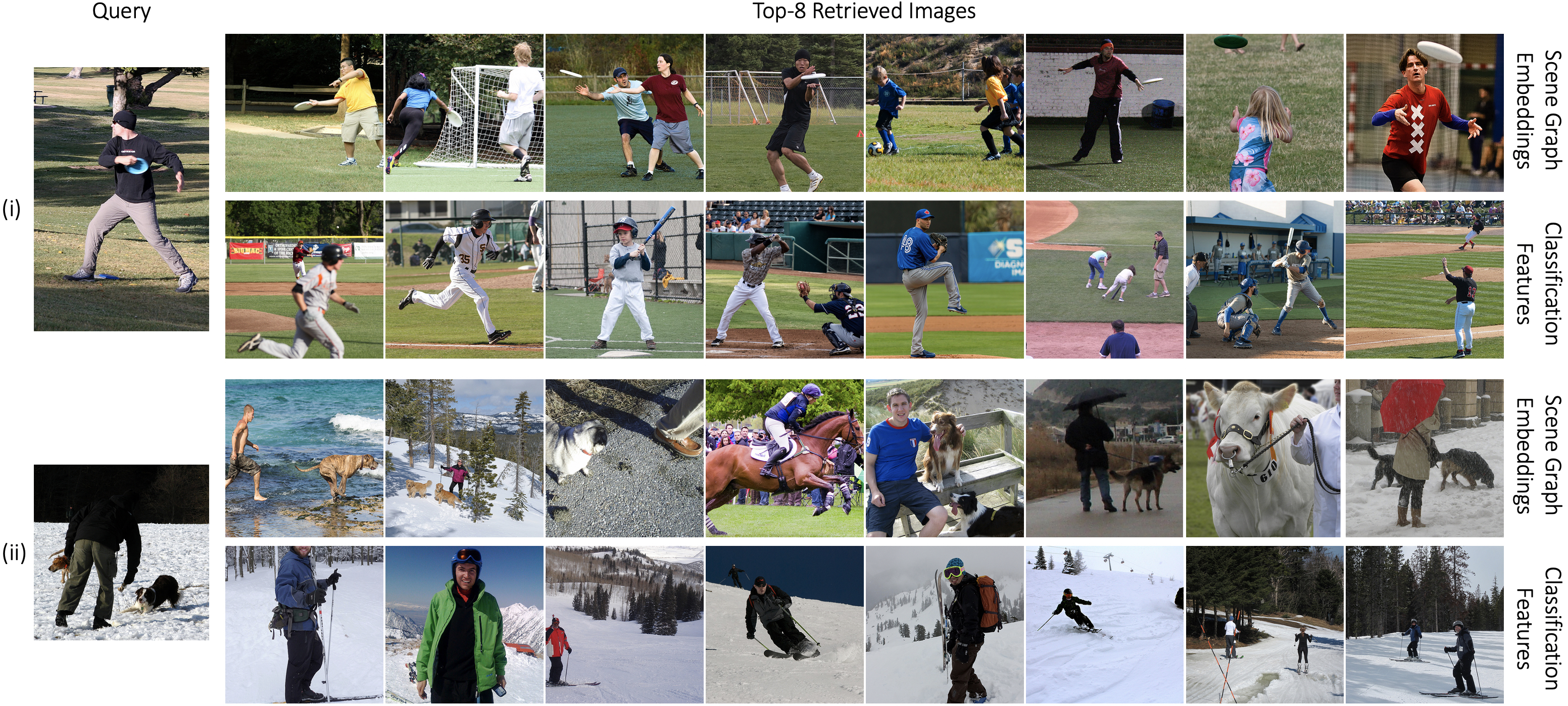}
    \caption{Comparison of image retrieval results for exemplar queries using scene graph embeddings from our proposed model, and classification features of ResNet-152~\cite{resnet}. While the latter retrieves visually similar images, our embeddings capture the global structure of the scene, i.e, `man throwing frisbee' and `man playing with dog' respectively. Notice that classification features do not distinguish various outdoor sports in (i), and fail to depict the human-animal interaction in (ii).}
    \label{ResnetComparison}
\end{figure*}

In recent times, the advancement of deep convolutional neural networks (CNNs) has led to significant improvements in image classification~\cite{resnet,krizhevsky2012imagenet}. Their intermediate image representations have also proven to be powerful visual descriptors in a variety of other tasks. One such application is content-based image retrieval, where a given image is used to query a database and retrieve other similar images. Models trained for image classification capture visually discriminative features and the retrieved images are, therefore, visually similar to the query image. On the other hand, semantic image retrieval enables the search of \textit{visual situations} where multiple objects interact in different ways such as `man talking on the phone while sipping coffee'. This requires an understanding of the semantics of the scene as well as bridging the \textit{semantic gap}~\cite{smeulders2000content,hare2006mind} between visual features and high-level concepts. To address this, numerous visual-semantic embedding models have been proposed that incorporate semantic information from the text modality into image representations. While natural language enables rich and detailed descriptions of visual content, it is unstructured and requires explicit grounding into images~\cite{krishnamurthy2013jointly,lin2014visual}. This has led to the recent development of image representations using graph-based formulations that capture detailed scene semantics in a structured format.

Scene graphs~\cite{johnson2015image} have emerged as one such popular and powerful representation of the underlying content of images. This construct encapsulates the constituent objects and their relationships, and also encodes object attributes and spatial information. Their success can be ascribed to the numerous downstream applications that they facilitate, including visual question answering~\cite{teney2017graph,norcliffe2018learning}, scene classification~\cite{schroeder2019triplet}, image manipulation~\cite{Dhamo_2020_CVPR} and visual relationship detection~\cite{lu2016visual}.

In the context of image retrieval, there has been work on grounding scene graphs into images to obtain the likelihood of the scene graph-image pair~\cite{johnson2015image,schuster2015generating}. Alternately, we propose to utilize distributed representations derived from scene graphs of images alongside standard measures of similarity such as cosine similarity or inner product. Embeddings derived from the scene graphs capture the information present in the scene and this allows us to combine the advantages of structured representations like graphs and continuous intermediate representations. Further, we demonstrate in Figure~\ref{ResnetComparison} that similarity search over these embeddings captures the overall context of the scene, offering an alternative to visual similarity provided by traditional image embeddings.

We leverage Graph Convolutional Networks to map an image's scene graph into an embedding. In the existing literature, the training of such models is set up in one of the following paradigms: ($1$)~Self-supervised, where scene graphs are jointly embedded with corresponding supplementary information (like visual features or text) about the image~\cite{wang2016learning,belilovsky2017joint}, and ($2$)~Task-dependent, where learning of the scene graph representation is driven by supervision from specific downstream applications~\cite{teney2017graph,schroeder2019triplet}. In contrast, we consider a pairwise similarity matrix as our supervision signal where every value represents a noisy notion of similarity between the corresponding image pair. We do not make any assumptions on this similarity (for example, it may not obey properties of a metric) and hence argue that our supervision is less strict. In this work, we define these similarities using the text modality, specifically image captions, but it can also be derived from other sources.

We show how robust representations can be learnt from scene graphs by leveraging caption similarities in a ranking context. This is enabled by a novel loss function that extracts signal from pairs of similar and dissimilar images, as in the contrastive learning approach. Contrary to the commonly-used class labels per image, we introduce soft target labels based on relative similarities to appropriately weigh the extent of similarity or dissimilarity. Further, we show improved retrieval performance with the learnt representations.

We summarize our key contributions as follows:
\begin{enumerate}
    \item A graph convolutional network to process scene graphs into a visual-semantic embedding space for images, which combines benefits of both the structured information present in graphs and distributed representations.
    \item A novel ranking loss that incorporates relative ranking constraints and outperforms other contrastive learning losses. Furthermore, a comparative analysis of different triplet sampling strategies is presented.
    \item The embeddings are demonstrated to be robust to noise via a retrieval experiment with incomplete scene graph queries. We also provide qualitative insights into the use of these representations for content-based image retrieval.
\end{enumerate}

\section{Related Work}
The relevant literature is introduced from the following two aspects: ($1$)~Scene Graphs and their Applications ($2$)~Visual-Semantic Embedding Models.

\subsection{Scene Graphs and their Applications}
The availability of large image datasets with detailed graph representations~\cite{krishna2017visual,lu2016visual} has led to a surge in research on scene graphs. A large fraction of this work focuses on generating scene graphs from images~\cite{xu2020survey}. They have also proven to be effective in a range of visual tasks such as image retrieval~\cite{johnson2015image,wang2020cross}, image generation~\cite{johnson2018image} and captioning~\cite{yang2019auto}.

There has been some work on learning scene graph representations for downstream applications. \citeauthor{raboh2020differentiable} propose intermediate representations, called differentiable scene graphs, that can be trained end-to-end with supervision from a reasoning task. \citeauthor{schroeder2019triplet} introduce triplet supervision and data augmentation to learn scene graph embeddings for layout generation. Distinct from their work, we aim to learn semantic embeddings for images using scene graphs, leveraging the text modality for weak supervision rather than specific downstream tasks.

The closest related work~\cite{belilovsky2017joint} considers joint representations of scene graphs and images by aligning scene graphs with pre-trained image features using neural networks. However, we do not wish to match scene graphs to visual features as they are object-centric and do not attend to the semantics of the scene. Our focus is to learn scene graph embeddings that capture the global context. 

Another line of relevant work involves image-text matching using graph structures. \citeauthor{li2019visual} propose a reasoning model to identify the key objects and relationships of a scene. \citeauthor{shi2019knowledge} incorporate common-sense knowledge from aggregate scene graphs to improve the matching process. \citeauthor{wang2020cross} take a more direct approach by developing graphs for both modalities and computing their similarity via custom distance functions. In this work, however, we leverage the text modality for weak supervision rather than embedding multiple modalities in the same space. In our retrieval setup, a scene graph query is used to run a nearest neighbor search over embeddings of the indexed images.

\subsection{Visual-Semantic Embedding Models}
Image representations obtained by deep convolutional networks have had tremendous success in a range of vision tasks. Early works~\cite{resnet,krizhevsky2012imagenet} focused on image classification using image-level labels. Follow up works include triplet formulations~\cite{wang2014learning} that produce more generally useful visual representations with reduced data requirements.

Some common directions to learn visual-semantic representations for images include the use of word embeddings of class labels~\cite{frome2013devise,li2017learning}, exploiting class structure for classification~\cite{yan2015hd} and leveraging WordNet ontology for class hierarchies~\cite{deng2011hierarchical,barz2019hierarchy}. These methods work for simple images but cannot be trivially extended to complex scenes with multiple objects and relationships.

More recent work considers the multimodal setting where pairwise ordering constraints are placed on both image and text modalities~\cite{kiros2014unifying,wang2016learning} in a ranking formulation for representation learning. Additionally, similarity networks~\cite{garcia2019learning} have been proposed that take as input a pair of images and train a network using regression objectives over pairwise similarity values. We build on the above in two directions: ($1$)~we derive embeddings from scene graphs (rather than pixel information or unstructured text) by utilizing a Graph Convolutional Network ($2$)~we leverage a weak pairwise similarity supervision from the text modality and design appropriate objective functions to drive the training of our model.

\section{Model}
We consider the task of learning image embeddings from a structured representation of its content. Each image $\mathcal{I}$ has a corresponding scene graph $\mathcal{G}_{\mathcal{I}} = (V_{\mathcal{I}}, E_{\mathcal{I}})$, where the vertices $V_{\mathcal{I}}$ represent objects and directed edges $E_{\mathcal{I}}$ denote the relationships between them. Therefore, $\mathcal{G}_{\mathcal{I}}$ comprises of $<$\textit{subject, predicate, object}$>$ triples such as $<$\textit{cat, on, bed}$>$ or $<$\textit{man, driving, car}$>$. We wish to learn a mapping $\Phi : \mathcal{G}_{\mathcal{I}} \rightarrow \mathbf{f}_{\mathcal{I}}$ where $\mathbf{f}_{\mathcal{I}} \in \mathbb{R}^D$ is the embedding of image $\mathcal{I}$. Inspired by recent work~\cite{johnson2018image} on learning intermediate scene graph representations, we model $\Phi$ as a Graph Convolutional Network (GCN). It performs a series of \textit{convolution} operations on the graph, followed by an aggregation layer to pool context from different entities in the image.

Each vertex $u$ and edge $e_{uv}$ is encoded as a vector, $\mathbf{\Lambda}_{u}~\in~\mathbb{R}^d$ and $\mathbf{\Lambda}_{uv}\in\mathbb{R}^d$ respectively, using separate learnable embedding layers. These vectors are updated by convolution operations from their respective immediate neighborhoods. For nodes, this update step is a function of all its one-hop neighbor nodes. And edge representations are updated based on the source and target node. Hence, the context is propagated throughout the graph via its edges. Mathematically, each convolutional layer in the GCN relays information across entities by applying the following operations in order:
\begin{enumerate}
    \item \textbf{Message Passing}: Each edge in the graph generates a ``message'' for its source and target nodes. For edge $e_{uv} \in E_{\mathcal{I}}$, a message $\mathbf{m}_{uv}^{s} \in \mathbb{R}^h$ is sent to the source node $u$ and another message $\mathbf{m}_{uv}^{t}\in \mathbb{R}^h$ is sent to the target node $v$. These messages gather information from the edge state $\mathbf{\Lambda}_{uv}$ and the node states $\mathbf{\Lambda}_{u}$ and $\mathbf{\Lambda}_{v}$ and are denoted by
    \begin{equation*}
        \mathbf{m}_{uv}^{s} \leftarrow \psi_s(\mathbf{\Lambda}_{u}, \mathbf{\Lambda}_{v}, \mathbf{\Lambda}_{uv})
    \end{equation*}
    \begin{equation*}
        \mathbf{m}_{uv}^{t} \leftarrow \psi_t(\mathbf{\Lambda}_{u}, \mathbf{\Lambda}_{v}, \mathbf{\Lambda}_{uv})
    \end{equation*}

    \item \textbf{State Update for Edges}: The state vector for an edge $\mathbf{\Lambda}_{uv}$ is updated to $\hat{\mathbf{\Lambda}}_{uv} \in \mathbb{R}^D $ by combining the most recent node states with the edge's prior state as
    \begin{equation*}
        \hat{\mathbf{\Lambda}}_{uv} \leftarrow \psi_{e} (\mathbf{\Lambda}_{u}, \mathbf{\Lambda}_{uv}, \mathbf{\Lambda}_{v})
    \end{equation*}

    \item \textbf{State Update for Nodes}: The state for every node $\mathbf{\Lambda}_{u}$ is updated to an intermediate representation which is obtained by pooling all the messages it receives via its edges
    \begin{equation*}   
        \mathbf{\Gamma}_{u} \leftarrow \cfrac{\sum_{w \mid (u,w)  \in E_{\mathcal{I}}} \mathbf{m}_{uw}^{s} + \sum_{w \mid (w,u) \in E_{\mathcal{I}}} \mathbf{m}_{wu}^{t}}{\sum_{w \mid (u,w) \in E_{\mathcal{I}}} 1 + \sum_{w \mid (w,u) \in E_{\mathcal{I}}} 1}
    \end{equation*}

    This intermediate pooled representation is passed through another non-linear transformation and normalized to produce the updated node state $\hat{\mathbf{\Lambda}}_{u} \in \mathbb{R}^D$ as
    \begin{equation*}
        \hat{\mathbf{\Lambda}}_{u} \leftarrow \cfrac{\psi_{n}(\mathbf{\Gamma}_{u})}{\|\psi_{n}(\mathbf{\Gamma}_{u})\|_2}
    \end{equation*}

    $\ell_2$-normalization results in unit length vectors and has been shown to provide superior results~\cite{chen2020simple}.
\end{enumerate}

The state vectors $\mathbf{\Lambda}_{u}$ and $\mathbf{\Lambda}_{uv}$ are iteratively updated via a series of graph convolutional layers such that the resulting state vectors of nodes capture information from the entire graph. Finally, we define the embedding of the scene graph (and image) as the average over all learnt node state vectors
\begin{equation*}
    \mathbf{f}_{\mathcal{I}} \leftarrow \cfrac{\sum_{u \in V_{\mathcal{I}}} \hat{\mathbf{\Lambda}}_{u}}{\sum_{u \in V_{\mathcal{I}}} 1}
\end{equation*}

All non-linear transformations -- $\psi_s, \psi_t, \psi_e, \psi_n$ -- are implemented as multilayer perceptrons. Specifically, the functions $\psi_s, \psi_t \text{ and } \psi_e $ are modeled using a single network that concatenates the inputs $\mathbf{\Lambda}_{u}, \mathbf{\Lambda}_{uv}, \mathbf{\Lambda}_{v}$ and computes $3$ outputs using separate fully connected heads. Weight sharing across all neighborhoods allows the layer to operate on graphs of arbitrary shapes. To ensure that a scene graph is connected, we augment a trivial node $\underline{\hspace{2mm}} image \underline{\hspace{2mm}}$ and trivial edges $\underline{\hspace{2mm}} in\_image \underline{\hspace{2mm}}$ from every other node to this node.

\subsection{Weak Supervision from the Text Modality}
The previous section described our GCN architecture that maps the scene graph for each image $\mathcal{A}$ in a collection of $N$ images into its corresponding embedding $\mathbf{f}_{\mathcal{A}}$. Our supervision signal for training the network is an $N\times N$ similarity matrix where entries $\mathrm{s}_{\mathcal{XY}}$ represent the measure of similarity between images $\mathcal{X}$ and $\mathcal{Y}$. In the current work, these similarities are computed using textual captions of corresponding images as natural language is key in conveying semantics. Further, single-sentence, user-generated captions tend to focus on the entirety of the scene.

\begin{figure}[!h]
    \centering
    \includegraphics[width=\columnwidth]{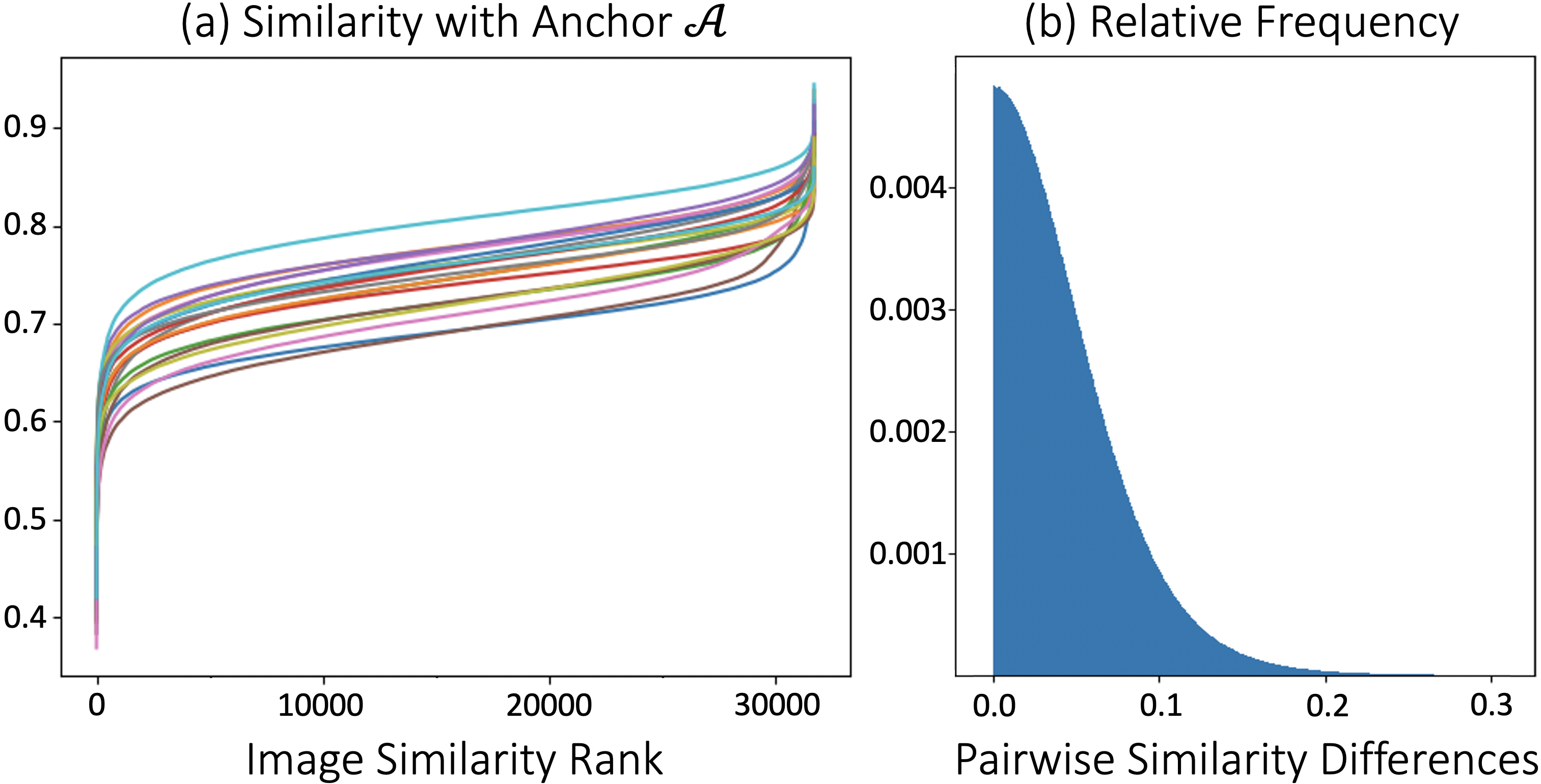}
    \caption{(a)~Each of the $20$ plots show the sorted similarities $\mathrm{s}_{\mathcal{AX}} \hspace{1mm} \forall \mathcal{X}$ for randomly chosen anchor images $\mathcal{A}$ (b)~Relative frequencies of the absolute values of all pairwise similarity differences $|\mathrm{s}_{\mathcal{AX}} - \mathrm{s}_{\mathcal{AY}}| \hspace{1mm} \forall \mathcal{X,Y}$ of the $20$ selected anchors.}
    \label{SimilarityPDFs}
\end{figure}

A strict criterion would be to set $\mathrm{sim}(\mathbf{f}_{\mathcal{X}},\mathbf{f}_{\mathcal{Y}}) \approx \mathrm{s}_{\mathcal{XY}}$. Our work follows similar lines, but instead of treating the similarities $\mathrm{s}_{\mathcal{XY}}$ as direct regression targets, we employ a contrastive approach to impose only ordering or ranking constraints. Our approach is motivated by the nature of the data, shown in Figure~\ref{SimilarityPDFs}a, where each curve corresponds to the sorted similarity values $\mathrm{s}_{\mathcal{AX}}$ of all images $\mathcal{X}$ with respect to an anchor image $\mathcal{A}$. We observe that the image captions are mostly equally distant from each other - represented by the narrow range from $0.6$ to $0.8$ in the middle of the plots. This is also corroborated in Figure~\ref{SimilarityPDFs}b where the distribution of absolute differences in similarity peaks at $0$ and steadily declines with the $99^\text{th}$ percentile occurring at $0.16$. Thus, learning embeddings $\mathbf{f}_{*}$ with regression objectives using Siamese or Triplet architectures~\cite{chopra2005learning} are likely to lead to degenerate solutions. 

We rely on the text modality to only provide weak supervision, i.e., we expect that the image scene graphs contain complementary information, with the text captions only providing a guiding signal for the model's training. To this end, we impose a lenient requirement that $\mathrm{sim}(\mathbf{f}_{\mathcal{A}},\mathbf{f}_{\mathcal{P}})~>~\mathrm{sim}(\mathbf{f}_{\mathcal{A}},\mathbf{f}_{\mathcal{N}})~\text{if}~\mathrm{s}_{\mathcal{AP}}~>~\mathrm{s}_{\mathcal{AN}}$. This formulation invokes a set of three images $\langle \mathcal{A}, \mathcal{P}, \mathcal{N} \rangle$ similar to well-known losses in contrastive learning. However, we need to account for the fact that the similarity of a \textit{positive} image $\mathcal{P}$ with respect to the \textit{anchor} $\mathcal{A}$ might be very close to that of \textit{negative} image $\mathcal{N}$. That is, $\mathrm{s}_{\mathcal{AP}}$ and $\mathrm{s}_{\mathcal{AN}}$ might occupy very similar regions in the density plot of $\mathrm{s}_{\mathcal{A*}}$. We therefore design a loss function that is tolerant to the selection of such samples during training.

\begin{figure*}[!h]
    \centering
    \includegraphics[width=\linewidth]{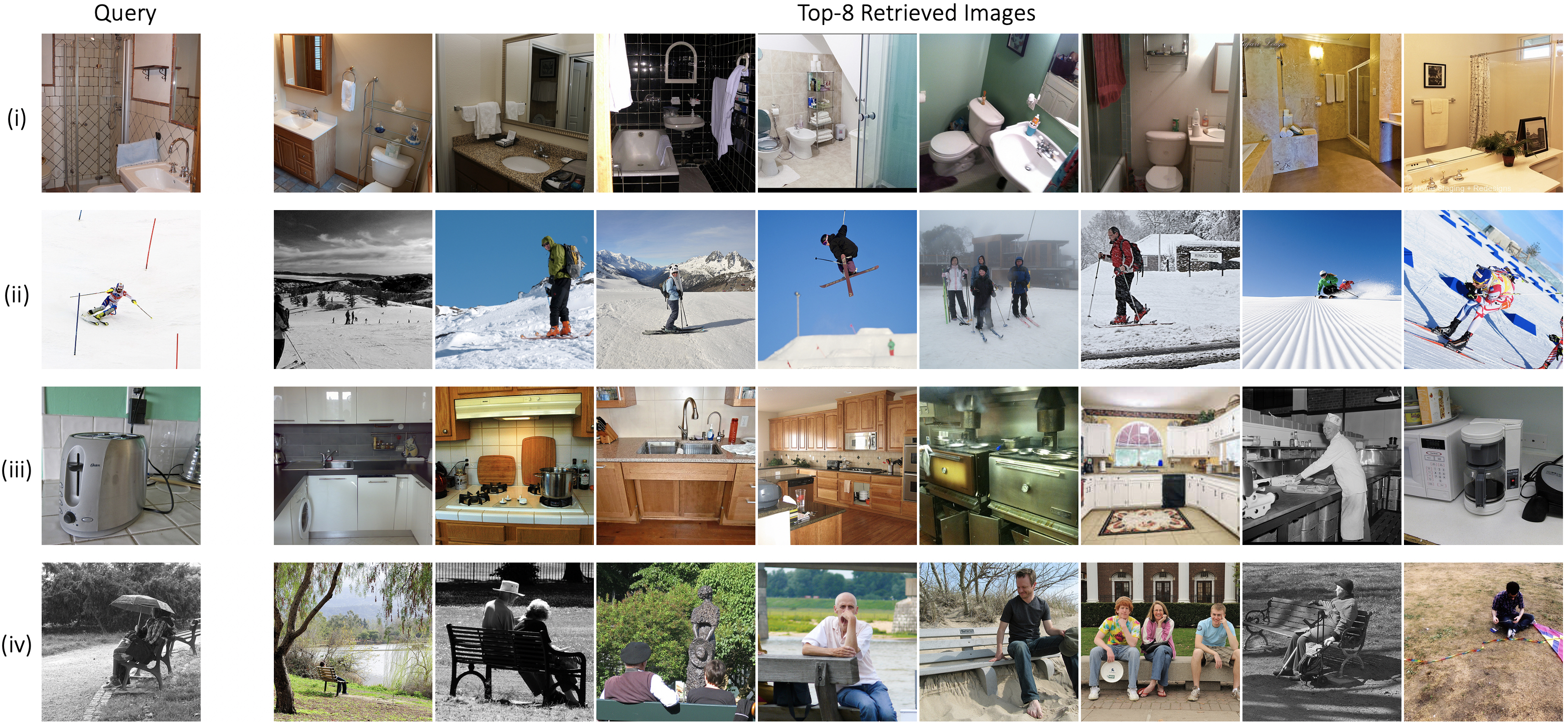}
    \caption{Qualitative examples showing Top-8 retrieved images of a given query using the proposed scene graph embeddings. Notice how the model is able to retrieve images of a kitchen in (iii) while the query image only contains a toaster. This can be attributed to the GCN framework which captures object co-occurrences in the scene graphs. In example (iv), the embeddings capture the global context of the query image - people sitting on a bench in an outdoor setting - while providing visual diversity.}
    \label{SimilarImages}
\end{figure*}

\subsection{Ranking Loss for Similarity Supervision}
In this section, we describe a novel loss function to learn image embeddings that is specifically designed to optimize our model given a continuous space of similarities (or distances) between images, rather than discrete labels which is common in classification literature. Inspired by RankNet~\cite{burges2005learning}, we model the posterior probability $\mathrm{\hat{P}}$ of having similarities in the correct order as
\begin{equation*}
    \hat{\mathrm{P}}(\hspace{1mm}\mathbf{f}_{\mathcal{A}}^{T}\mathbf{f}_\mathcal{P}^{} > \mathbf{f}_{\mathcal{A}}^{T}\mathbf{f}_\mathcal{N}^{}\hspace{1mm}) = \sigma \left( \frac{\mathbf{f}_{\mathcal{A}}^{T}\mathbf{f}_\mathcal{P}^{}-\mathbf{f}_{\mathcal{A}}^{T}\mathbf{f}_\mathcal{N}^{}}{\nu} \right)
\end{equation*}
where $\sigma$ is the sigmoid function, $\nu$ is a temperature hyperparameter and we have utilized the inner product $\mathbf{f}_{\mathcal{A}}^{T}\mathbf{f}_\mathcal{X}^{}$ for the similarity function $\mathrm{sim}(\mathbf{f}_{\mathcal{A}},\mathbf{f}_{\mathcal{X}})$. For a given anchor $\mathcal{A}$, $\mathcal{P}$~(positive) and $\mathcal{N}$~(negative) are such that the pair $(\mathcal{A}, \mathcal{P})$ are expected to be more similar than $(\mathcal{A}, \mathcal{N})$. Since the corresponding embeddings $\mathbf{f}_{*}$ are $\ell_2$-normalized, the inner products above correspond to using cosine similarity.

To reflect the constraints from relative similarities, we define the desired target value $\mathrm{P}$ as
\begin{equation*}
    \mathrm{P}(\hspace{1mm}\mathrm{s}_{\mathcal{AP}} > \mathrm{s}_{\mathcal{AN}}\hspace{1mm}) = \frac{\mathrm{s}_{\mathcal{AP}}}{\mathrm{s}_{\mathcal{AP}} + \mathrm{s}_{\mathcal{AN}}}
\end{equation*}
where $\mathrm{s}_{\mathcal{AP}}~\text{and}~\mathrm{s}_{\mathcal{AN}}$ denote the caption similarity of the anchor with the positive and negative respectively. If both negative and positive are sampled with high confidence such that $\mathrm{s}_{\mathcal{AP}} \gg \mathrm{s}_{\mathcal{AN}}$, then $\mathrm{P} \approx 1$. However, as explained in Figure~\ref{SimilarityPDFs}, such samples are uncommon in our dataset. The proposed setup is efficient as it allows the use of samples where $\mathrm{s}_{\mathcal{AP}}$ is only marginally more than $\mathrm{s}_{\mathcal{AN}}$ with an appropriately weighted contribution to the objective. Hence, the use of non-binary targets offers an alternative to the explicit mining of positive and negative samples.

The final loss function, termed as Ranking loss hereafter, takes the form of a cross entropy and is given by
\begin{equation*}
    \mathcal{L} = - \mathrm{P} \log \hat{\mathrm{P}} - (1-\mathrm{P})\log \hspace{1mm} (1-\hat{\mathrm{P}})
\end{equation*}

Optimizing this loss function allows us to learn an embedding space in which the similarity between scene graph embeddings respects the ordering or ranking indicated in the similarity matrix. The advantages of this setup are: ($1$)~the similarity values $\mathrm{s}_{\mathcal{A}*}$ are not assumed to be transitive or obey triangle inequalities, and ($2$)~the actual magnitude of the similarities are not part of the supervision, only the relative values. Therefore, the Ranking loss imposes minimal requirements from the supervision signal.

\begin{table}[!h]
\centering
\fontsize{9.0pt}{10.0pt}\selectfont
\begin{tabular}{cc}
\toprule
\textbf{Name} & \textbf{Loss Function} \\
\midrule
\addlinespace
Triplet &  $\max\left(\hspace{1mm}\mathbf{f}_{\mathcal{A}}^{T}\mathbf{f}_\mathcal{N}^{} - \mathbf{f}_{\mathcal{A}}^{T}\mathbf{f}_\mathcal{P}^{} + m, 0\hspace{1mm}\right)$\\
\addlinespace
\addlinespace
InfoNCE &  $-\log \dfrac{\exp\left(\hspace{0.5mm}\mathbf{f}_{\mathcal{A}}^{T}\mathbf{f}_\mathcal{P}^{}/\lambda\right)}{\exp\left(\hspace{0.5mm}\mathbf{f}_{\mathcal{A}}^{T}\mathbf{f}_\mathcal{P}^{}/\lambda\right)+\exp\left(\hspace{0.5mm}\mathbf{f}_{\mathcal{A}}^{T}\mathbf{f}_\mathcal{N}^{}/\lambda\right)}$  \\
\addlinespace
\bottomrule
\end{tabular}
\caption{Loss functions used in contrastive learning}
\label{lossFunctions}
\end{table}

We compare the proposed loss function with other commonly used losses in the contrastive learning paradigm, namely Triplet or Margin loss~\cite{weinberger2006distance,schroff2015facenet} and InfoNCE~\cite{oord2018representation}, shown in Table~\ref{lossFunctions}. Note that contrastive learning as defined by ~\citeauthor{chen2020simple} is equivalent to our formulation if: ($i$) Class labels are used to set $s_{\mathcal{AX}}=1$ if the two images $\mathcal{A}$ \& $\mathcal{X}$ belong to the same class and $0$ otherwise ($ii$) The~$\langle \mathcal{A}, \mathcal{P}, \mathcal{N} \rangle$~triples are chosen such that $s_{\mathcal{AP}}=1$ and $s_{\mathcal{AN}}=0$. We believe that our setup is naturally more robust to alternate ways of sampling the triples, and we experiment with different options described next.

\begin{table*}[!h]
\centering
\fontsize{9.0pt}{10.0pt}\selectfont
\begin{tabular}{lrc ccc c ccc}
\toprule
\multicolumn{2}{c}{\textbf{Model}} && \multicolumn{3}{c}{\textbf{Per Image (Row-wise) Evaluation}} && \multicolumn{3}{c}{\textbf{All Pairs Evaluation}} \\
\cmidrule{1-2}\cmidrule{4-6}\cmidrule{8-10}
\textbf{Objective} & \textbf{Sampling} && \textbf{Kendall $\tau$} & \textbf{Spearman $\rho$} & \textbf{Pearson $r$} && \textbf{Kendall $\tau$} & \textbf{Spearman $\rho$} & \textbf{Pearson $r$}\\
\midrule
\multirow{4}{*}{Triplet} & Random && $0.258$ & $0.375$ & $0.402^{ \dagger}$ && $0.251$ & $0.369$ & $0.389^{ \dagger}$ \\
 & Extreme && $0.133$ & $0.197$ & $0.264$ && $0.148$ & $0.220$ & $0.274$ \\
 & Probability && $0.235$ & $0.345$ & $0.358$ && $0.224$ & $0.333$ & $0.335$ \\
 & Reject && $0.269^{ \dagger}$ & $0.392^{ \dagger}$ & $0.388$ && $0.254^{ \dagger}$ & $0.375^{ \dagger}$ & $0.364$ \\[0.1ex]
\hdashline\noalign{\vskip 0.6ex}
\multirow{4}{*}{InfoNCE} & Random && $0.263^{ \dagger}$ & $0.382^{ \dagger}$ & $0.407^{ \dagger}$ && $0.253^{ \dagger}$ & $0.372^{ \dagger}$ & $0.388^{ \dagger}$ \\
 & Extreme && $0.034$ & $0.048$ & $0.049$ && $0.062$ & $0.081$ & $0.081$ \\
 & Probability && $0.187$ & $0.275$ & $0.336$ && $0.197$ & $0.290$ & $0.333$ \\
 & Reject && $0.225$ & $0.329$ & $0.376$ && $0.228$ & $0.337$ & $0.367$ \\[0.1ex]
\hdashline\noalign{\vskip 0.6ex}
\multirow{4}{*}{Ranking} & Random && $0.381$ & $0.539$ & $0.548$ && $0.366$ & $0.524$ & $0.525$ \\
 & Extreme && $0.207$ & $0.302$ & $0.361$ && $0.209$ & $0.309$ & $0.359$ \\
 & Probability && $\bf{0.391}^{ \dagger}$ & $\bf{0.554}^{ \dagger}$ & $\bf{0.549}^{ \dagger}$ && $\bf{0.377}^{ \dagger}$ & $\bf{0.540}^{ \dagger}$ & $\bf{0.529}^{ \dagger}$ \\
 & Reject && $0.380$ & $0.537$ & $0.538$ && $0.362$ & $0.520$ & $0.511$ \\[0.1ex]
\hdashline\noalign{\vskip 0.6ex}
\multicolumn{2}{c}{Normal Features} && $0.003$ & $0.004$ & $0.024$ && $0.001$ & $0.002$ & $0.017$ \\
\multicolumn{2}{c}{Classification Features} && $0.260$ & $0.378$ & $0.417$ && $0.247$ & $0.363$ & $0.382$ \\
\bottomrule
\end{tabular}
\caption{Evaluation of the models trained using different objective functions and sampling techniques. \textit{Normal Features} denotes embeddings drawn at random from a normal distribution and \textit{Classification Features} are pre-trained visual embeddings~\cite{resnet}. The best results are highlighted in boldface. ${ }^\dagger$ indicates the best results for a given objective function.}
\label{trainingResults}
\end{table*}

\subsection{Sampling Techniques}
We provide different strategies to sample a positive $\mathcal{P}$ and negative $\mathcal{N}$ for a given anchor image $\mathcal{A}$. We do this by leveraging the caption similarities $\mathrm{s}_{\mathcal{AX}}$ of the anchor $\mathcal{A}$ with every other image $\mathcal{X}$. The sampling alternatives are:
\begin{enumerate}
    \item \textbf{Random}: Given an anchor $\mathcal{A}$, we sample uniformly at random a positive-negative pair $\langle \mathcal{P},\mathcal{N} \rangle$ from the set of all correctly-ordered pairs given by
    \begin{equation*}
        \{\hspace{1mm} \langle \mathcal{P}' , \mathcal{N}' \rangle \hspace{2mm}|\hspace{2mm}  \mathrm{s}_{\mathcal{AP}'} > \mathrm{s}_{\mathcal{AN}'} \hspace{0.5mm}\}
    \end{equation*}
    While this ensures that the positive is closer to the anchor than the negative, it does not consider the relative distances between them.
    \item \textbf{Extreme}: For every anchor image $\mathcal{A}$, we pick the most similar image as the positive $\mathcal{P}$ and the most dissimilar image as the negative $\mathcal{N}$. Mathematically
    \begin{equation*}
        \hspace{0.5mm}\mathcal{P}\hspace{0.5mm} = \argmax_{\mathcal{P}'} \hspace{1mm} \mathrm{s}_{\mathcal{AP}'} \quad\quad \hspace{0.5mm}\mathcal{N}\hspace{0.5mm} = \argmin_{\mathcal{N}'} \hspace{1mm} \mathrm{s}_{\mathcal{AN}'}
    \end{equation*}
    Note that this is a deterministic method, i.e., same positive and negative examples for a given anchor across epochs.
    \item \textbf{Probability}: We sample the positive and negative based on their caption similarities with the anchor as
    \begin{equation*}
        \mathrm{P} (\hspace{0.5mm}\mathcal{P}\hspace{0.5mm}) = \dfrac{\mathrm{s}_{\mathcal{AP}}}{\sum\limits_{\mathcal{P}'}\mathrm{s}_{\mathcal{AP}'}} \quad\quad \mathrm{P} (\hspace{0.5mm}\mathcal{N}\hspace{0.5mm}) = \dfrac{1 - \mathrm{s}_{\mathcal{AN}}}{\sum\limits_{\mathcal{N}'} (1 - \mathrm{s}_{\mathcal{AN}'} ) }
    \end{equation*}
    The upper limit of caption similarities $\mathrm{s}_{\mathcal{AX}}$ is $1$ and therefore, $1-\mathrm{s}_{\mathcal{AX}}$ is a measure of distance between images $\mathcal{A}$ and $\mathcal{X}$. This sampling captures the intuition that images closer (farther) to the anchor should have a higher probability of being sampled as the positive (negative).
    \item \textbf{Reject}: Though infrequent, sampling based on similarities might lead to cases where $\mathcal{N}$ is closer to $\mathcal{A}$ than $\mathcal{P}$. In this method, we follow Probability sampling with an added constraint on the direction of pairwise similarities, i.e., rejecting samples where $\mathrm{s}_{\mathcal{AP}} < \mathrm{s}_{\mathcal{AN}}$.
\end{enumerate}

The loss functions in Table~\ref{lossFunctions} typically utilize strategies where \textit{hard negatives} are coupled with a positive~\cite{Wang_2015_ICCV}, or conversely \textit{easy positives} alongside negatives~\cite{levi2020reducing} to aid the learning. The Ranking loss and sampling techniques described are designed to leverage as many of the $N^2$ positive-negative pairs for a given anchor as possible. Given the observed benefits of multiple negatives~\cite{chen2020simple} and multiple positives~\cite{khosla2020supervised} per anchor, future work will look into adapting the methods above to handle the exhaustive range of triples.

\section{Experiments}
We describe the dataset, experimental setup and both qualitative and quantitative evaluation of our approach.

\textbf{Dataset}: We work on the Visual Genome~\cite{krishna2017visual} dataset which is a collection of $108,077$ images and their scene graphs. We use a subset of $51,498$ images which have a set of $5$ user-provided textual captions in MS-COCO~\cite{mscoco}. We also filter object and relationship types that occur at least $25$ times, resulting in $2416$ object and $478$ relationship categories. We use images with minimum $3$ and maximum $40$ objects, and at least one relationship. This results in $45,358$ images with an average of $21$ objects and $15$ relationships per image. We divide the data into train, validation and test sets with a $70:20:10$ split. Further, we consider the $5$ captions available for each image and embed them by taking the average of the constituent word embeddings~\cite{glove}. The image similarities $\mathrm{s}_{\mathcal{XY}}$ are defined as the average of the $5 \times 5 = 25$ pairwise inner products over caption embeddings.

\textbf{Implementation Details}: Both objects and relationships are first embedded into a $d=300$ dimensional space using separate learnable layers. These are initialized with the average of constituent GloVe embeddings~\cite{glove}. The intermediate messages for nodes are $h=512$ size vectors, while the final node and edge states of each layer are $D=300$ size vectors. For all multilayer perceptrons, we use ReLU activation and batch normalization~\cite{batchnorm}. The model consists of $5$ GCN layers and is trained using Adam optimizer~\cite{adam} for $100$ epochs with learning rate $10^{-4}$ and batch size $16$. The temperature parameter in InfoNCE and Ranking loss has been set as $\lambda=1$ and $\nu=1$ and the margin in Triplet loss as $m=0.5$. Training is performed on a Ubuntu 16.01 machine, using a single Tesla V100 GPU and PyTorch framework.

\begin{figure*}[!h]
    \centering
    \includegraphics[width=\linewidth]{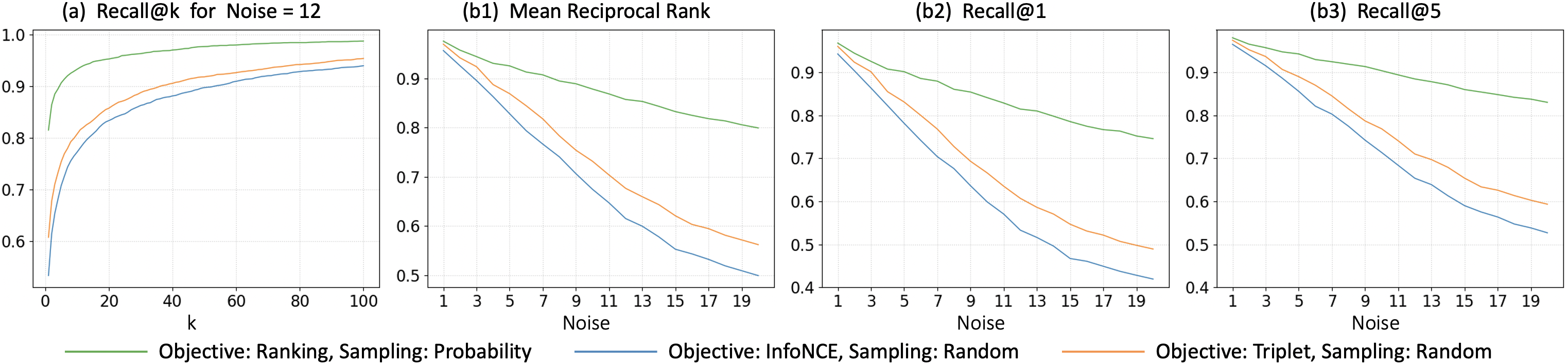}
    \caption{Retrieval performance for models trained on different objectives and corresponding best sampling: (a) Recall@k versus k for noise = $12$ (b) Variation of the ranking metrics with increasing noise levels, from $1$ to $20$, in the query scene graph.}
    \label{RetrievalPlots}
\end{figure*}

\textbf{Evaluation}: We compute the GCN output $\mathbf{f}_{\mathcal{X}}$ for every image $X$ in the test set and derive the pairwise similarities as $\mathrm{sim}(\mathbf{f}_{\mathcal{X}},\mathbf{f}_{\mathcal{Y}}) = \mathbf{f}_{\mathcal{X}}^{T}\mathbf{f}_\mathcal{Y}^{}$. These scene graph similarities are compared against the corresponding caption equivalents $\mathrm{s}_{\mathcal{XY}}$ using Kendall rank correlation coefficient $\tau$, Spearman's rank correlation coefficient $\rho$ and Pearson correlation coefficient $r$. The $2$ rank correlations are of primary importance as our model is trained on relative similarities, and not absolute values. We compute the metrics at two levels - per image (or row-wise) and across all pairs. The micro-averaged option of computing row-wise correlation between model-derived scene graph similarities $\mathrm{sim}(\mathbf{f}_{\mathcal{X}},\mathbf{f}_{\mathcal{*}})$ and caption similarities $\mathrm{s}_{\mathcal{X*}}$ reflects the retrieval focus in this work.

The results are tabulated in Table~\ref{trainingResults} and we make the following observations:
(a) Comparing across the $3$ objective functions, it is evident that the proposed Ranking loss consistently outperforms the Triplet and InfoNCE alternatives for any sampling. The magnitude of improvements is particularly noticeable in the \textit{All Pairs Evaluation}.
(b) Amongst the $4$ sampling methods, Random is a robust alternative across loss functions, while the deterministic sampling strategy, Extreme, performs the worst. Comparing between Probability and Reject sampling - the Triplet and InfoNCE losses that utilize binary labels perform better when coupled with Reject strategy. In Ranking loss, however, Probability sampling outperforms as it is based on soft target labels and can handle invalid triples (when the positive is further away from the anchor than the negative).
(c) The best performing model, trained using Ranking loss and Probability sampling combination, has a Kendall $\tau$ of $0.391$. Note that a perfect value of $1$ would be undesirable as it indicates that the scene graph modality contains redundant information with respect to the textual captions.
(d) The classification features~\cite{resnet} perform very competitively. This is particularly impressive given that they are pre-trained and not customized to the current task. The qualitative comparison in Figure~\ref{ResnetComparison} provides some intuition on how the two embeddings differ. Further, Figure~\ref{SimilarImages} shows nearest neighbors of query images and indicates the viability of the scene graph embeddings.

\section{Image Retrieval}
We design a retrieval setup to highlight the robustness of our scene graph embeddings to missing objects and relationships. We believe that our model trained over observed scene graphs produces embeddings that implicitly contain contextual information about objects that tend to co-occur and their relationships. To illustrate this, we consider every image in the test set and introduce increasing levels of noise in the scene graphs. We eliminate a set of $M$ edges chosen at random from the scene graph and subsequently drop all isolated objects which are disconnected from the rest of the graph. This \textit{noisy scene graph} is passed through the GCN model to get a query embedding which is issued against the test set. We examine the ranked list of items and evaluate the model's ability to retrieve the known image. The objective and sampling combination where the relevant image continues to be returned in top ranks despite noisy input is deemed to be most immune to incomplete information in the query.

The results for noise level $M=12$ (median number of edges across scene graphs) are shown in Table~\ref{retrievalTable}. We compute the retrieval performance using standard metrics - Mean Reciprocal Rank (MRR), Recall@1 and Recall@5. It can be seen that Ranking loss generates embeddings that are significantly more effective than Triplet and InfoNCE losses. In fact, even the worst performing model of Ranking loss outperforms all variants of the other two losses. The best combination is the Ranking loss alongside the Probability sampling based on similarities. The target image is returned in top-$5$ ranks (out of $4537$ in test set) in over $90\%$ of the cases. The increased levels of recall are observed even as we go further down the ranked list, as shown in Figure~\ref{RetrievalPlots}(a).

\begin{table}[!h]
\fontsize{9.0pt}{10.0pt}\selectfont
\centering
\begin{tabular}{lr c lll }
\toprule
\multicolumn{2}{c}{\textbf{Model}} && \multicolumn{3}{c}{\textbf{Retrieval Metrics}} \\
\cmidrule{1-2}\cmidrule{4-6}
\textbf{Objective} & \textbf{Sampling} && \textbf{MRR} & \textbf{R@1} & \textbf{R@5} \\
\midrule
\multirow{4}{*}{Triplet} & Random && $0.676^{ \dagger}$ & $0.607^{ \dagger}$ & $0.753^{ \dagger}$ \\
 & Extreme && $0.431$ & $0.370$ & $0.496$ \\
 & Probability && $0.208$ & $0.138$ & $0.265$ \\
 & Reject && $0.337$ & $0.242$ & $0.431$ \\[0.1ex]
\hdashline\noalign{\vskip 0.6ex}
\multirow{4}{*}{InfoNCE} & Random && $0.615^{ \dagger}$ & $0.533^{ \dagger}$ & $0.707^{ \dagger}$ \\
 & Extreme && $0.005$ & $0.002$ & $0.005$ \\
 & Probability && $0.145$ & $0.094$ & $0.185$ \\
 & Reject && $0.576$ & $0.502$ & $0.655$ \\[0.1ex]
\hdashline\noalign{\vskip 0.6ex}
\multirow{4}{*}{Ranking} & Random && $0.710$ & $0.643$ & $0.789$ \\
 & Extreme && $0.700$ & $0.648$ & $0.758$ \\
 & Probability && $\bf{0.857}^{ \dagger}$ & $\bf{0.815}^{ \dagger}$ & $\bf{0.906}^{ \dagger}$ \\
 & Reject && $0.712$ & $0.639$ & $0.793$ \\
\bottomrule
\end{tabular}
\caption{Retrieval performance for different models. Queries are noisy scene graphs (noise = $12$) and images are ranked on their similarities in the embedding space. The best results are in boldface. ${ }^\dagger$ indicates the best results for a given objective.}
\label{retrievalTable}
\end{table}

For generalization, we sweep the noise parameter by progressively removing a chosen number of edges (and isolated objects) from the query scene graph up to a maximum of $20$ edges ($3^{rd}$ quartile for the number of edges across scene graphs). We compute the same metrics as before but restrict our attention to the best sampling strategy for each objective (marked with ${ }^\dagger$ in Table~\ref{retrievalTable}). The results are provided in Figure~\ref{RetrievalPlots}(b). It can be observed that the Ranking loss alongside Probability sampling performs the best across all three metrics. Despite removing $75\%$ of edges in the query scene graph, a nearest neighbor search in the proposed embedding space places the target image at rank $1$ in over $70\%$ of the cases. This indicates the robustness of our scene graph representations and validates the model setup - a graph convolutional network to compute image embeddings, and the novel ranking loss that effectively utilizes pairwise similarity constraints as a weak supervision signal.

\section{Conclusion and Discussion}
We have considered the setting of retrieval of images based on their scene content. To do this, we obtained embeddings from ground-truth scene graphs of images using a graph convolutional network. This model was trained using a weak supervision signal of pairwise similarity preferences obtained from the text modality. We have proposed a loss function based on relative similarity labels, and have shown superior performance of the derived embeddings in a retrieval task. There are at least $2$ promising future directions: ($1$)~leveraging progress in scene graph generation literature allows the representation learning method to be applicable in a wide set of scenarios ($2$)~the training objective based on weaker supervision requirements is general and relevant in other situations where classification-based labels are not available.

\bibliography{references}

\end{document}